\pdfoutput=1 
\documentclass{article}

\usepackage[final,nonatbib]{neurips_2021}

\usepackage[utf8]{inputenc} 
\usepackage[T1]{fontenc}    
\usepackage{hyperref}       
\usepackage{url}            
\usepackage{booktabs}       
\usepackage{amsfonts}       
\usepackage{nicefrac}       
\usepackage{microtype}      
\usepackage{xcolor}         
\usepackage{graphicx}
\usepackage{bm}
\usepackage{algorithm}
\usepackage{algorithmic}
\usepackage{amsmath}
\usepackage{multirow}

\title{3rd Place Solution for VisDA 2021 Challenge --

Universally Domain Adaptive Image Recognition}

\author{%
  Haojin Liao\textsuperscript{1,5} ~~~~ Xiaolin Song\textsuperscript{2,5}  ~~~~ Sicheng Zhao\textsuperscript{3} ~~~~ Shanghang Zhang\textsuperscript{4}  ~~~~ Xiangyu Yue\textsuperscript{4}
  \\ \textbf{Xingxu Yao\textsuperscript{5}} ~~~~ \textbf{Yueming Zhang\textsuperscript{2,5}} ~~~~ \textbf{Tengfei Xing\textsuperscript{5}} ~~~~ \textbf{Pengfei Xu\textsuperscript{5}} ~~~~ \textbf{Qiang Wang\textsuperscript{1}}\\
  1. Beijing University of Posts and Telecommunications ~~~~~~~~  2. Tianjin University\\3. Columbia University ~~~~~~~~ 4. UC Berkeley ~~~~~~~~5. Didi Chuxing\\
  \texttt{liaohaojin@bupt.edu.cn} \\
}

\begin{document}

\maketitle

\begin{abstract}
  The Visual Domain Adaptation (VisDA) 2021 Challenge calls for unsupervised domain adaptation (UDA) methods that can deal with both input distribution shift and label set variance between the source and target domains. In this report, we introduce a universal domain adaptation (UniDA) method by aggregating several popular feature extraction and domain adaptation schemes. First, we utilize VOLO, a Transformer-based architecture with state-of-the-art performance in several visual tasks, as the backbone to extract effective feature representations. Second, we modify the open-set classifier of OVANet to recognize the unknown class with competitive accuracy and robustness. As shown in the leaderboard, our proposed UniDA method ranks the 3rd place with 48.49\% ACC and 70.8\% AUROC in the VisDA 2021 Challenge.
\end{abstract}

\section{Introduction}

In real-world applications, there might be no large quantities of labeled data to train a task model, especially a deep neural network. Domain adaptation (DA) aims to learn a model from other labeled source domains that can generalize and adapt to novel target distributions~\cite{zhao2020review}. This year's Visual Domain Adaptation Challenge (VisDA 2021) focuses on an open-world setting. On the one hand, the input distribution is different between the source and target domains. On the other hand, the label set varies without prior knowledge, i.e., there might be missing and/or novel classes in the target data as in the Universal Domain Adaptation (UniDA) setting~\cite{saito2021ovanet}. It is challenging for a DA method to simultaneously obtain high accuracy of known classes and robust performance of the missing and/or novel classes. In our solution to the VisDA 2021 challenge, we aggregate several popular modules and schemes for feature extraction and universal domain adaptation. Inspired by the success of Transformer architectures in various computer vision tasks, especially the image recognition task, we choose VOLO~\cite{yuan2021volo} as the backbone for effective feature representation. To deal with the missing or unknown classes in the target domain, we follow OVANet~\cite{saito2021ovanet}, a state-of-the-art UniDA method. Differently, we modified main parts of the model and introduced a new backbone and an adversarial domain discriminator to enhance the robustness of the model. Out of the 24 submissions on the leaderboard, our solution ranks the 3rd place. 


\section{Our Solution}
\begin{figure*}[!ht]
    \centering
    \includegraphics[width=1.0\textwidth]{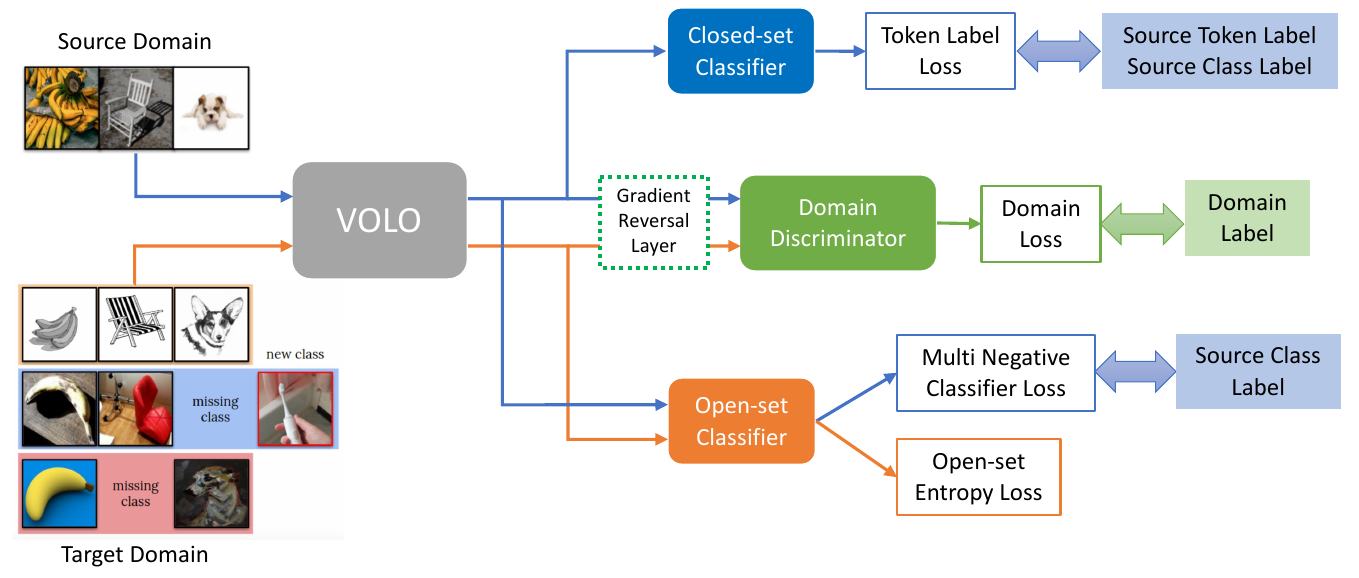}
    \caption{Overview of our proposed solution with four main parts that cooperate with each other. First, the backbone VOLO is employed to extract features of both source and target images. And then these features are sent to the closed-set classifier, open-set Classifier, and domain discriminator, separately. The closed-set classifier is used to identify a possible known class while the open-set Classifier is used to determine whether the sample is known or unknown. Domain discriminator helps the backbone to match the feature distributions of the source and target data from adversarial training for the known class.}
    \label{overview}
\end{figure*}

In this section, we will introduce the proposed solution in detail: (1) the adopted source model in Section 2.1 and (2) the employed universal domain adaptation strategy in Section 2.2., which includes the baseline method and its defects, closed-set classifier, and open-set classifier. An overview of our proposed solution is shown as Figure~\ref{overview}.

\subsection{Source Model: VOLO}
Effective feature representations play an important role in domain adaptation. In the UDA setting, we can train a model only on the source data, the performance of which can reflect the effectiveness of feature representations. Since Vision Transformer (ViT)~\cite{dosovitskiy2020image} was proposed, Transformer architecture has become more and more popular in computer vision. However, many versions of Transformer need to be trained on a large-scale dataset to obtain good performance. Meanwhile, some kinds of Transformer can still obtain competitive classification accuracy on ImageNet~\cite{ILSVRC15} without extra training data, such DeiT~\cite{touvron2021training}, BEiT~\cite{bao2021beit}, and VOLO~\cite{yuan2021volo}.
According to the model size and external data requirements in the VisDA-2021 challenge, i.e., 100 million parameters is limited in total and no other external data is allowed, we select VOLO-D3 which obtains 86.3\% top1 classification accuracy with 86 million parameters on ImageNet without other training data.

\subsection{Universal Domain Adaptation}

\paragraph{Baseline method} 
We employ OVANet~\cite{saito2021ovanet} as the baseline UniDA method, which achieves the state-of-the-art performance on the OfficeHome, VisDA, and DomainNet datasets. There are three main components in OVANet. The first one is a feature extractor that aims to extract effective feature representations for both the source and target data in a shared space. The second one is a closed-set classifier that is trained on the labeled source data to learn the mappings between feature representations and known categories. The last one is an open-set classifier consisting of a bunch of binary classifiers trained on both source and target data, which helps to align known target samples to source samples and keep unknown instances as unknown. Although OVANet achieves good adaptation performance on OfficeHome, VisDA, and DomainNet datasets, it dose not perform well on ImageNet. This is probably because that the number of categories on ImageNet is much larger. 
In our solution, we propose to improve performance of OVANet from two aspects. One is to replace the feature extractor with one that works better on ImageNet and the other is to improve the training and domain adaptation methods.

\paragraph{Closed-set classifier}
All VOLO's pretrained parameters are kept when used in OVANet. The feature extractor of OVANet is replaced with the backbone of VOLO and the closed-set classifier is replaced with VOLO's head. In this way, we can learn a more effective classifier from the supervision of labeled source data. However, when following the original training settings of OVANet, the accuracy drops while the AUROC is the same. As we know, Transformer-based models are usually difficult to train, which motivates us to follow VOLO's training strategies. The most important one is using the Token Labeling method~\cite{jiang2021all} for data augmentation. The input of VOLO is the patches of an image and the output is the feature for each patch. If only the class label is used as the supervised information, the detailed information of the patches will be lost. Token Labeling is one method to provide supervised information for each patch of an image. Following VOLO, we employ CutOut\cite{zhong2020random} and RandAug\cite{cubuk2020randaugment} removing brightness and contrast which are on the blacklist of the challenge as data augmentation. Please note that the data augmentation methods are only applied on the source data.


\paragraph{Open-set classifier on source data}
In OVANet, the open-set classifier is composed of a bunch of binary classifiers, the number of which is the same with the number of categories. During the training of OVANet, there are only two binary classifiers producing loss in each training step. One is the positive classifier and the other one is the nearest negative classifier. This can be problematic when the number of categories is large, because there might be many near negative classifiers. Using only the nearest one makes it difficult for the model to converge. To deal with this issue, we increase the number of near negative binary classifiers that would produce loss. 

\paragraph{Open-set classifier on target data}
In order to align the known target samples to source samples, OVANet computes entropy of all the binary classifiers and takes the average as the loss to train the model. Although entropy measures the uncertainty of a system, OVANet considers each binary classifier as a single system when computing entropy. In other words, as long as every binary classifier makes a sharp prediction, we can obtain a very low entropy loss and the uncertainty of the whole open-set classifier cannot be well discriminated. By this way, the model can increase the confidence of unknown class, but it has little help to improve the accuracy of known classes' target samples. To deal with this, we introduce an adversarial domain discriminator in the network. Like UAN~\cite{you2019universal}, the domain discriminator can help match the feature distributions of the source and target data adversarially for the known class. By adding a gradient reversal layer between the feature extractor and domain discriminator, domain-invariant features can be obtained and the accuracy on the target domain can be improved.

\section{Experiments}

\subsection{Dataset and Evaluation}

\paragraph{Dataset}
In VisDA 2021 challenge, 
the annotated ImageNet dataset is used as the source data, and the target domains use images from the following public datasets -- ObjectNet, ImageNet-R, ImageNet-C, and ImageNet-O. Besides, under the UDA setting, the labels of source data are allowed to train the model but the target labels are unknown. 

\paragraph{Evaluation metrics}
The final adaptation result is evaluated by a combination of two metrics: classification accuracy (ACC) and area under the ROC curve (AUROC). ACC is calculated only on the known categories, while the AUROC measures unknown category detection by thresholding a score that represents how likely the input belongs to an unknown class.

\subsection{Implementation Details}
The detailed implementation of VOLO and and the pretrained modal on ImageNet without using any extra training data are available at \url{https://github.com/sail-sg/volo}. In order to incorporate the token labeling method into our training, the labeled token data on ImageNet should be first generated which can be download at \url{https://github.com/zihangJiang/TokenLabeling}. Our model is trained on a machine with 4 Tesla-P40 GPUs. The batch size is 64 which means that there are 128 samples from both source and target domains sent to the model on each step. Because of the lack of GPUs, our model is only trained on images with a size of 224x224. The final model consists of all the methods mentioned in the previous sections and is trained by a two-stage approach. 

In the first stage, we utilize the AdamW optimizer with an original learning rate of 8.0e-6 for the open-set classifier and adversarial domain discriminator and 4.0e-6 for the feature extractor and closed-set classifier. Cosine learning rate scheduler is used to decay initial learning rate with a warm-up strategy. For the open-set classifier, we choose 300 near negative binary classifiers to produce loss. The loss of the adversarial domain discriminator is:
\begin{equation}
    L_d = -\mathbb{E}_{x\sim p}\log D(F(x)) - \mathbb{E}_{x\sim q}w^t(x)\log (1 - D(F(x)))
\end{equation}
where ${x\sim p}$ and ${x\sim q}$ respectively indicate that sample $x$ belongs to the source domain $p$ and target domain $q$, and $w^t(x)$ indicates the probability of a target sample $x$ belonging to the unknown class which is generated by the open-set classifier. The model is trained for 30,000 steps in the first stage.

In the second stage, the adversarial domain discriminator is removed and the optimizer is changed to momentum SGD with an original learning rate of 1.0e-3 for the open-set classifier and 8.0e-6 for the feature extractor and closed-set classifier. This helps the model to improve the ability of recognizing the unknown images. The model is trained for 8,000 steps in the second stage. We utilize the trained model from the second stage model for inference like the original OVANet. The difference is that we first crop out the top left, top right, bottom left, bottom right, and the middle for each test image and then send these five crops to our model. Finally, we compute the mean values of the output for the five crops.


\begin{table}[]
\caption{Ablation study on the validation set.}
\label{ablation-study}
\centering
\begin{tabular}{ccccccccc}
\toprule
\multicolumn{1}{c}{Method}       & \multicolumn{8}{c}{Performance}                               \\
\midrule
Source Model (VOLO-D3) & $\checkmark$ & $\checkmark$ & $\checkmark$ & $\checkmark$ & $\checkmark$ & $\checkmark$ & $\checkmark$ & $\checkmark$ \\
Data Augmentation &  & $\checkmark$ & $\checkmark$ & $\checkmark$ & $\checkmark$ & $\checkmark$ & $\checkmark$ & $\checkmark$ \\
Token Labeling &  &  &  & $\checkmark$ & $\checkmark$ & $\checkmark$ & $\checkmark$ & $\checkmark$ \\
Near Classifiers-300 &  &  & $\checkmark$ & $\checkmark$ & $\checkmark$ & $\checkmark$ &  & $\checkmark$ \\
Domain Discriminator  &  &  &  &  & $\checkmark$ & $\checkmark$ & $\checkmark$ & $\checkmark$ \\
2-Stage Training &  &  &  &  &  & $\checkmark$ & $\checkmark$ & $\checkmark$ \\
5-Crops Inference &  &  &  &  &  &  & $\checkmark$ & $\checkmark$ \\
\midrule
ACC         & 53.28 & 50.12 & 47.89 & 53.31 & 60.14 & 59.79 & 58.69 &59.93 \\
AUROC      & 60.02 & 66.61 & 69.91 & 67.65 & 59.28 & 71.63 & 70.92 &72.06 \\
\bottomrule
\end{tabular}
\end{table}

\begin{table}[]
\caption{Leaderboard in VisDA 2021 Challenge. The top-5 methods are listed.}
\label{leaderboard}
\centering
\begin{tabular}{lccccc}
\toprule
\multicolumn{1}{c}{\multirow{2}{*}{Rank}} & \multicolumn{1}{c}{\multirow{2}{*}{User}} & \multicolumn{2}{l}{Adapted Model} & \multicolumn{2}{l}{Source Model} \\ 
\multicolumn{1}{c}{} & \multicolumn{1}{c}{} & ACC & AUROC & ACC & AUROC \\ 
\midrule
1	&babychick & 56.29 &	69.79 &	56.29  & 69.79  \\
2   &chamorajg  & 48.49 &	76.86 &	0.07 &	50.00  \\
3	&\textbf{liaohaojin (Ours)} & 48.49 & 70.8 & 41.25 & 64.48  \\
4	&DXM-DI-AI-CV-TEAM &48.60 &	68.29 &	25.70 &62.43 \\
5   &fomenxiaoseng & 45.23 & 78.76 & 40.22 &	60.43 \\
\bottomrule
\end{tabular}
\end{table}

\subsection{Ablation Study and Comparison Results}

As shown in Table~\ref{ablation-study}, we report the results of different method combinations on the validation set. First, we shows the performance of the source model VOLO-D3.``Near Classifiers-300'' means that in open-set classifier there are 300 near negative binary classifiers that would produce loss. Then we use token labeling and adversarial domain discriminator method to train the model which significantly improve the accuracy while the 2-stage training method improves the ability of identifying unknown samples. We obtain the final result by 5-crops inference with 59.93\% ACC and 72.06\% AUROC while the source only model can produce 53.28\% ACC and 60.02\% AUC on validation set.

In the NeurIPS2021 VisDA challenge, our proposed domain adaptation framework ranks the 3rd place on the leaderboard with 48.49\% ACC and 70.8\% AUROC. Performance of top five teams are shown in Table~\ref{leaderboard}.


\small
\bibliographystyle{unsrt} 
\bibliography{neurips21.bib}
\end{document}